\documentclass[11pt]{elsarticle}

\let\today\relax
\makeatletter
\def\ps@pprintTitle{%
    \let\@oddhead\@empty
    \let\@evenhead\@empty
    \def\@oddfoot{\footnotesize\itshape
         {} \hfill\today}%
    \let\@evenfoot\@oddfoot
    }
\makeatother

\usepackage{hhline}
\usepackage{booktabs}
\usepackage{setspace}
\usepackage{amsfonts}
\usepackage{amssymb}
\usepackage{algpseudocode}
\usepackage{mathrsfs}
\usepackage{multirow}
\usepackage{tikz}
\usepackage{setspace}
\usepackage{etoolbox}   
\usepackage{lipsum}
\usepackage{booktabs}
\usepackage{graphicx,wrapfig}
\usepackage{mathtools}
\usepackage{comment}
\usepackage[hyphens]{url}
\usepackage{booktabs}

\RequirePackage{amsmath,amsfonts,amssymb}
\RequirePackage{natbib}

\usepackage{bbm}
\usepackage{enumitem}
\usepackage{graphicx}
\usepackage{nicefrac}
\usepackage{url}
\usepackage{algorithm}
\usepackage{multirow}
\usepackage{color}
\usepackage{arydshln}
\newcommand{\Pb}{\hbox{{I}\kern-.1667em\hbox{P}}} 
\newcommand\myatop[2]{\genfrac{}{}{0pt}{}{#1\hfill}{#2\hfill}}

\AtBeginEnvironment{tabular}{\doublespacing}

\usetikzlibrary{arrows,automata,backgrounds,fit,patterns,petri,shadows,shapes, positioning}

\usepackage[top=3.5cm,bottom=3.5cm,left=3.5cm,right=3.5cm]{geometry}

\usepackage{amsthm}
\newtheorem{property}{Property}

\usepackage{lineno}
\usepackage[hidelinks]{hyperref}
\modulolinenumbers[5]

\usepackage{arydshln}
\usepackage{amssymb}
\usepackage{mathtools}
\usepackage{mathrsfs}  
\usepackage{float}
\usepackage{graphicx}  

\usepackage{amsmath}
\usepackage{graphicx,psfrag,epsf}
\usepackage{enumerate}
\usepackage{natbib}
\usepackage{comment}
\usepackage[official]{eurosym}
\usepackage{eurosym}
\usepackage{url} 
\usepackage{siunitx}
\usepackage{bbm}

\usepackage{url}            
\usepackage{booktabs}       
\usepackage{subcaption}
\usepackage{amsfonts}       
\usepackage{nicefrac}       
\usepackage{microtype}      

\usepackage{amsmath}
\usepackage{mathabx}
\usepackage{amsopn}

\usepackage{algpseudocode}
\usepackage{algorithm}

\algrenewcommand\algorithmicindent{0.5em}%

\makeatletter
\renewcommand{\Function}[2]{%
  \csname ALG@cmd@\ALG@L @Function\endcsname{#1}{#2}%
  \def\jayden@currentfunction{#1}%
}

\newcommand{\funclabel}[1]{%
  \@bsphack
  \protected@write\@auxout{}{%
    \string\newlabel{#1}{{\jayden@currentfunction}{\thepage}}%
  }%
  \@esphack
}

\usepackage{color}
\usepackage{caption}

\usepackage{float}
\floatstyle{plaintop}
\restylefloat{table}

\usepackage{amsthm}
\usepackage{bm}

\DeclareMathOperator*{\argmax}{arg\,max}

\usepackage{color,soul}
\usepackage[usestackEOL]{stackengine}[2013-09-11]

\setstackgap{L}{1.4\baselineskip}
\usepackage{tikz}
\usetikzlibrary{arrows,automata,backgrounds,fit,patterns,petri,shadows,shapes, positioning}
\pgfdeclarepatternformonly{stripes}
{\pgfpointorigin}{\pgfpoint{0.25cm}{0.25cm}}
{\pgfpoint{0.25cm}{0.25cm}}
{
    \pgfpathmoveto{\pgfpoint{0cm}{0cm}}
    \pgfpathlineto{\pgfpoint{0.25cm}{0.25cm}}
    \pgfpathlineto{\pgfpoint{0.25cm}{0.12cm}}
    \pgfpathlineto{\pgfpoint{0.12cm}{0cm}}
    \pgfpathclose%
    \pgfusepath{fill}
    \pgfpathmoveto{\pgfpoint{0cm}{0.12cm}}
    \pgfpathlineto{\pgfpoint{0cm}{0.25cm}}
    \pgfpathlineto{\pgfpoint{0.12cm}{0.25cm}}
    \pgfpathclose%
    \pgfusepath{fill}
}

\usepackage{tabularx}
\usepackage{booktabs}

\begin{document}

\begin{frontmatter}

\title{Bayesian Graph Traversal}


\author[my1address]{William N. Caballero}
\author[my1address]{Phillip R. Jenkins}
\author[duke]{David Banks}
\author[my1address]{Matthew Robbins}

\address[my1address]{Department of Operational Sciences, Air Force Institute of Technology, WPAFB, OH}
\address[duke]{Department of Statistical Science, Duke University, Durham, NC}

\begin{abstract}
This research considers Bayesian decision-analytic approaches toward the traversal of an uncertain graph. Namely, a traveler progresses over a graph in which rewards are gained upon a 
node's first visit and costs are incurred for every edge traversal. 
The traveler knows the graph's adjacency matrix and his starting position, 
but does not know the rewards and costs. 
The traveler is a Bayesian who encodes his beliefs about these values using a Gaussian process prior and who seeks to maximize his expected utility over these beliefs. Adopting a decision-analytic perspective, we develop sequential decision-making solution strategies for this 
coupled information-collection and network-routing problem. 
We show that the problem is NP-Hard, and derive properties of the optimal walk. 
These properties provide heuristics for the traveler's problem that balance 
exploration and exploitation.
We provide a practical case study focused on the use of unmanned aerial systems for public safety and empirically study policy performance in myriad Erd\"os–R\'enyi settings.
\end{abstract}


\end{frontmatter}

\section{Introduction}
\label{sec:intro}

A traveler knows the adjacency matrix of a graph and the vertex at which he is located.
Each time an edge is traversed, there is a cost.
When a new node is reached, there is a reward.
The costs and rewards are unknown, but the traveler is a Bayesian with a joint prior distribution over them. 
If the traveler backtracks over an edge, the previous cost is assessed, but node rewards are
only received once.
Adopting decision-analytic principles, the traveler stops when the future expected
utility is negative or a fixed number of decisions has been made. 


This class of problems is quite general and captures many situations in sequential 
decision making.
One has prior beliefs, makes a choice, observes the results, updates the prior,
and then makes another decision.
Such a structure, as well as the network-traversal application, relates it to the Canadian Traveler Problem \citep{nikolova2008route, shiri2019randomized} as well as the {\color{black} uncertain vehicle-routing} work of \citet{han2014robust} and {\color{black} the research of \citet{borrero2016sequential} on shortest-path interdiction with incomplete information}, among others. However, the generality of our problem links it to many settings beyond transportation, e.g., choosing a career path, portfolio management, and staged investments \citep{lim2006sequential, summers2021friction}. This feature is due to the formulation embedding an information collection problem within a routing problem.

Ideally, the traveler's policy would map all locations and belief states to an 
optimal action. 
{\color{black}This ideal requires examination of all possible walks (i.e., a connected sequence of nodes) that start at his current location} and
determining their effects on his posterior beliefs given the observed costs and rewards. 
But, as we will see, such a solution is intractable. 
Even if edge costs and node rewards were known with certainty (i.e., perfect information is available), the problem is NP-hard. 
Therefore, we describe strategies that scale to larger networks and which approximate the 
optimal solution.

Related work in \citet{ryzhov2011information} solves a similar problem but assumes 
learning occurs before the traveler's route selection (i.e., offline). 
It distinguishes the information collection and decision phases with emphasis on the 
efficacy of the knowledge-gradient policy in the first phase.
\citet{ryzhov2012information} expand this work to general linear programming problems. 
In contrast, this paper addresses online learning, in which the information collection and 
decision phases are intertwined; the traveler learns about the graph while interacting with it. {\color{black}This perspective aligns with that of \citet{thul2023information}, which examines the routing of drones for forest-fire mitigation efforts.} 

Our work is also related to other stochastic network problems 
\citep[e.g., see][]{cohen2021minimax, min2022learning}; {\color{black} a comprehensive review of such problems is provided by \citet{powell1995stochastic}. Particularly relevant is the body of research on the stochastic vehicle routing problems \citep[e.g., see][]{secomandi2001rollout, bent2004scenario, soeffker2022stochastic}, as well as more recent work on the online shortest path \citep[e.g., see][]{lagos2024online} and the information-collection vehicle routing problems \citep[e.g., see][]{al2023information}. Although related, our underlying} modeling framework is derived from the partially observed Markov decision 
processes of \citet{aksakalli2016based} as well as the belief state
representation of \citet{powell2021reinforcement}. 
Moreover, we note that our work is adjacent to typical Bayesian optimization approaches in its use of Gaussian process priors 
\citep{shahriari2015taking, frazier2018tutorial}; {\color{black}however, it diverges  
since information collection is constrained by the graph's structure, thereby paralleling, to a degree, \textit{exotic} Bayesian optimization problems \citep[e.g., see][]{gardner2014bayesian}.} 

In Section \ref{sec:ProbForm}, we define the problem using the universal framework of
\citet{powell2019unified}; we also discuss assumptions on the traveler's prior. 
In Section \ref{sec:Insights}, we set aside the information collection stage to derive the
computational difficulties a clairvoyant traveler faces even with perfect information. 
This analysis gives insight into the structure of the traveler's optimal walk. 
Using this analysis, Section \ref{secAlgs} provides a collection of policies the traveler may 
use to solve the problem. 
These policies are tested empirically in Section \ref{sec:CompResults}. 
Section \ref{sec:Conclusion} summarizes our findings.



\section{The Traveler's Problem} \label{sec:ProbForm}

The traveler's known, undirected graph is $\mathcal{G}=(\mathcal{N},\mathcal{E})$, with
$\mathcal{N}$ the node set and $\mathcal{E}$ the edge set.
Edge-traversal costs are incurred every time an edge is crossed {\color{black} in either direction}; node rewards accrue
only on the first visit.
{\color{black}Both are unknown, but we assume that they are 
functions of known covariates, $\mathbf{x}_e \in \mathbb{R}^M$, $e\in\mathcal{E}$ and $\mathbf{y}_i \in \mathbb{R}^N$, $i\in\mathcal{N}$, where $M$ indicates the number of edge features and $N$ indicates the number of node features.}
The edge-traversal costs are an unknown function $c(\mathbf{x}_e)$, and the node rewards are 
an unknown function $r(\mathbf{y}_i)$. 
We assume that each node has a self-edge with 
zero cost traversal; this allows travel to stop under a fixed time horizon. 
Also, we alias $(i,j)$ and $(j, i)$, using both terms to refer to the same edge. Utility is assumed to accumulate additively based upon the difference between reward and cost; therefore, 
to maintain focus on the underlying dynamics, we often refer to utility as \textit{net gain}.    

Under these conditions, the traveler wants to identify an expected-utility-maximizing walk 
through the graph from some $i_0 \in \mathcal{N}$. 
However, due to the unknown functions $c(\mathbf{x}_e)$ and $r(\mathbf{y}_i)$, 
the traveler's decisions are coupled with his subjective beliefs. 
Since the traveler learns about these functions while traversing the network, 
his decisions depend on both his location and belief state, implying that the 
traveler's behavior is best determined by a policy. 
Assuming independent Gaussian process priors over $c$ and $r$, 
we present a tractable formulation of the traveler's problem using the  
framework in \citet{powell2019unified}. 
Policies for this problem are given in Section \ref{secAlgs}. 


\subsection{The Traveler's Prior Beliefs} \label{sec:TravelersPrior}

We assume {\color{black}independent} $c\sim \mathcal{GP}(\mu_c, \Sigma_c)$ and $r\sim \mathcal{GP}(\mu_r, \Sigma_r)$ where 
$\mu_c$ and $\mu_r$ are mean functions from $\mathbbm{R}^M\to\mathbbm{R}$ and 
$\mathbbm{R}^N\to\mathbbm{R}$, respectively, and $\Sigma_c$ and $\Sigma_r$ are 
their corresponding covariance functions. 
For a Gaussian process, any finite collection of its unknown realizations has a 
multivariate normal distribution. 
Also, the posterior of a Gaussian process given an observation is also a Gaussian process. 
These properties enable computation of the traveler's posterior beliefs given 
observed costs and rewards. 

\sloppy Consider the prior distribution over $k$ unobserved edge-traversal costs; denote the 
sequences of edges as $e_{1:k}=(e_1,\cdots, e_k)$. 
Then the prior over $c(\mathbf{x}_{e_{1:k}}) = [c(\mathbf{x}_{e_1}), \cdots, c(\mathbf{x}_{e_k})]$
is 
$$
c\left(\mathbf{x}_{e_{1:k}}\right)  \sim \text{Normal}\left(\mu_c\left(\mathbf{x}_{e_{1:k}}\right), 
\Sigma_c\left(\mathbf{x}_{e_{1:k}},\mathbf{x}_{e_{1:k}}\right) \right).
$$
We mirror the notation used by \citet{frazier2018bayesian} for a function applied to a
collection of points. That is, $\mathbf{x}_{e_{1:k}}$ indicates a sequence of feature vectors 
$\mathbf{x}_{e_1}, \cdots, \mathbf{x}_{e_k}$; $\mu_c(\mathbf{x}_{e_{1:k}}) = 
[\mu_c(\mathbf{x}_{e_1}), \cdots, \mu_c(\mathbf{x}_{e_k})]^\top$ is a mean vector, and 
$\Sigma_ c(\mathbf{x}_{e_{1:k}},\mathbf{x}_{e_{1:k}}) = 
[\Sigma_c(\mathbf{x}_{e_1},\mathbf{x}_{e_1}), \cdots, \Sigma_c(\mathbf{x}_{e_1},\mathbf{x}_{e_k}); 
\cdots ; \Sigma_c(\mathbf{x}_{e_k},\mathbf{x}_{e_1}), \cdots, 
\Sigma_c(\mathbf{x}_{e_k},\mathbf{x}_{e_k})]$ is a covariance matrix. 
Since $c(\mathbf{x}_{e_{1:k}})$ is multivariate normal, the posterior on 
$c\left(\mathbf{x}_{e_{k}}\right)$ given the data $\mathbf{x}_{e_{1:k-1}}$ is 
univariate normal with
\begin{align}
c\left(\mathbf{x}_{e_{k}}\right)  \sim 
\text{Normal}\left(\mu_{c}\left(\mathbf{x}_{e_k}\right),\sigma^2_{c }\left(\mathbf{x}_{e_k}\right)
\right) \label{eqPosterior}
\end{align}
where 
\begin{align*} 
\mu_{c}\left(\mathbf{x}_{e_k}\right) &= \Sigma_c \left(\mathbf{x}_{e_k}, 
\mathbf{x}_{e_{1:k-1}}\right) \Sigma_c 
\left(\mathbf{x}_{e_{1:k-1}}, \mathbf{x}_{e_{1:k-1}}\right) ^{-1}\left(c(\mathbf{x}_{e_{1:k-1}}) - 
\mu_c(\mathbf{x}_{e_{1:k-1}}) \right) + \mu_c\left(\mathbf{x}_{e_k}\right), \\
\sigma^2_{c} \left(\mathbf{x}_{e_k}\right) &= \Sigma_c(\mathbf{x}_{e_k},\mathbf{x}_{e_k}) - 
\Sigma_c \left(\mathbf{x}_{e_k}, \mathbf{x}_{e_{1:k-1}}\right) \Sigma_c 
\left(\mathbf{x}_{e_{1:k-1}}, \mathbf{x}_{e_{1:k-1}}\right) ^{-1}\Sigma_c 
\left(\mathbf{x}_{e_{1:k-1}}, \mathbf{x}_{e_k}\right).
\end{align*}

The posterior for multiple unobserved edges, $e_{k:K}$, is
\begin{align}
c\left(\mathbf{x}_{e_{k:K}}\right)  \sim 
\text{Normal}\left(\mu_{c}\left(\mathbf{x}_{e_{k:K}}\right),\text{cov}_{c}
\left(\mathbf{x}_{e_{k:K}}\right) \right) 
\end{align}
where
\begin{align*} 
\mu_{c}\left(\mathbf{x}_{e_{k:K}}\right) &= \Sigma_c \left(\mathbf{x}_{e_{k:K}}, 
\mathbf{x}_{e_{1:k-1}}\right) \Sigma_c \left(\mathbf{x}_{e_{1:k-1}}, 
\mathbf{x}_{e_{1:k-1}}\right)^{-1}\left(c\left(\mathbf{x}_{e_{1:k-1}}\right) -
\mu_c\left(\mathbf{x}_{e_{1:k-1}}\right) \right) + \mu_c\left(\mathbf{x}_{e_{k:K}}\right), \\
\text{cov}_{c} \left(\mathbf{x}_{e_{k:K}}\right) &=
\Sigma_c\left(\mathbf{x}_{e_{k:K}},\mathbf{x}_{e_{k:K}}\right) - \Sigma_c 
\left(\mathbf{x}_{e_{k:K}}, \mathbf{x}_{e_{1:k-1}}\right) \Sigma_c \left(\mathbf{x}_{e_{1:k-1}}, 
\mathbf{x}_{e_{1:k-1}}\right)^{-1} \Sigma_c \left(\mathbf{x}_{e_{1:k-1}}, 
\mathbf{x}_{e_{k:K}}\right).
\end{align*}

Similarly, conditional on the payoffs from the observed nodes, the posterior on 
the unobserved nodes is 
\begin{align}
r\left(\mathbf{y}_{j_{l:L}}\right)  \sim \text{Normal}\left(\mu_{r} 
\left(\mathbf{y}_{j_{l:L}}\right), \text{cov}_{r}\left(\mathbf{y}_{j_{l:L}}\right) \right) 
\label{eqPostR}
\end{align}
where $j_{1:l}$ is a sequence of $l$ nodes, $\mathbf{y}_{j_{1:l-1}}$ and 
$r(\mathbf{y}_{j_{1:l-1}})$ are observed, and $\mu_{r}$ and $\text{cov}_{r}$ are 
analogously defined. 
We shall use these posterior distributions to solve the traveler's problem in 
Section \ref{secAlgs}.

Unlike other applications of Gaussian processes (e.g., Bayesian optimization), 
$c$ and $r$ cannot be queried at an arbitrary feature vector. 
The traveler can only observe the functions at a new feature vector if it corresponds 
to a node/edge adjacent to his current position. 
This fact, coupled with the traveler's objective to maximize his net gain, complicates 
the problem and distinguishes it from other knowledge acquisition settings. 

\subsection{Problem Formulation under a Gaussian Process Prior} \label{sec:AGen}

We present a formulation for the traveler's problem that simultaneously characterizes 
the knowledge acquisition and reward-accrual dynamics. 
It uses the universal framework of \citet{powell2019unified} for sequential 
decision making, which includes explicitly defined state variables, decision variables, 
exogenous information, a state-transition model, and an objective function.
There is a slight deviation from the standard notation in order to accommodate
conventions in graph theory and statistics.

Let $\mathcal{T} = \{0,1,\ldots, T\}$ denote the set of decision epochs. 
Each decision epoch $ t\in \mathcal{T}$ represents a particular point in time 
at which the traveler makes a decision.  


\subsubsection{State Variables}

State variables describe the traveler's location and beliefs at a given time.  
This information is necessary and sufficient to compute the traveler's net gain at time $t$ 
and to derive the transition  model that updates the information needed for decisions 
at time $t+1$. 

At some time $t$, the traveler must decide which edge to traverse to reach the next node,
or whether to terminate the exploration.
By equations \eqref{eqPosterior}--\eqref{eqPostR}, his beliefs at time $t$ are determined 
from all previously observed data, as well as his prior mean and covariance functions. 
Because the latter two functions characterize the traveler's prior beliefs, 
this implies that the state variables can be defined as the tuple
$$
S_t = \langle i_t, \mathcal{I}_t, \mathcal{C}_{t}, \mathcal{R}_{t}\rangle
$$
where $i_t$ is the traveler's current location, $\mathcal{I}_t$ lists the visited nodes, 
$\mathcal{C}_{t}$ contains all observed $\mathbf{x}_{e_k}$ and $c(\mathbf{x}_{e_k})$ pairs, 
and $\mathcal{R}_{t}$ contains all observed $\mathbf{y}_{j_l}$ and $r(\mathbf{y}_{j_l})$ pairs.

Different indices are used for $\mathbf{x}_{e_k}$ and $\mathbf{y}_{j_l}$ because, 
depending upon $\mathcal{I}_t$, not all sampled $\mathbf{x}$- or $\mathbf{y}$-values 
will yield a new observation. 
If node $j_t$ has already been visited then its $r(\mathbf{y}_{j_t})$ is already known. 
But if $(i_t,j_t)$ has not yet been traversed, observing it provides
new information on $c$.  

\subsubsection{Decision Variables}

At time $t$, the traveler must choose which node $j_t$ to visit from his current position $i_t$.
The traveler's choice $j_t$  must be in $\mathcal{N}(i_t) \cup \{i_t\}$ such that 
$\mathcal{N}(i_t)$ is the set of all nodes adjacent to $i_t$. 
If $j_t=i_t$, the traveler stays in place. 

\subsubsection{Exogenous Information} Given the decision $j_t$, the exogenous information is
determined by the covariates associated with both $j_t$ and $e_t = (i_t,j_t)$. 
Namely, if $j_t$ has not yet been visited, the traveler observes the values of the unknown
functions at this input, i.e.,  $c(\mathbf{x}_{e_t})$ and $r(\mathbf{y}_{j_t})$. 
Alternatively, if $j_t$ has been visited but $e_t$ has not been traversed, 
the traveler only sees $c(\mathbf{x}_{e_t})$ because no reward is received. 
If the traveler crosses a previously traversed edge, no new information is observed; 
the known travel cost for $c(\mathbf{x}_{e_t})$ accrues and no node reward is received. 

\subsubsection{Transition Model}
After having made decision $j_t$ and observing $c(\mathbf{x}_{e_t})$ and $r(\mathbf{y}_{j_t})$, the state variables update to enable the tractable modeling of the traveler's next decision. In this setting, the transition model simply involves moving the traveler to his new location and appending information to the relevant sets. Therefore, after making a decision $j_t$ and observing the exogenous information, the new state becomes

$$
S_{t+1} = \langle j_t, \mathcal{I}_{t+1}, \mathcal{C}_{t+1},  \mathcal{R}_{t+1} \rangle.
$$  

\noindent where 

\begin{align}
\mathcal{I}_{t+1} &= \begin{cases} \mathcal{I}_t \cup \{j_t\}, \quad &j_t \notin \mathcal{I}_t, \\
\mathcal{I}_t, \quad & j_t \in \mathcal{I}_t,\end{cases} \label{eq:I_t} \\ 
\mathcal{C}_{t+1} &= \begin{cases} \mathcal{C}_{t} \cup \left\{\left(\mathbf{x}_{e_t}, c(\mathbf{x}_{e_t})\right)\right\}, \quad & \left(\mathbf{x}_{e_t}, c(\mathbf{x}_{e_t})\right) \notin \mathcal{C}_{t}, \\
\mathcal{C}_{t}, \quad & \left(\mathbf{x}_{e_t}, c(\mathbf{x}_{e_t})\right) \in \mathcal{C}_{t},\end{cases}\\
\mathcal{R}_{t+1} &= \begin{cases} \mathcal{R}_{t} \cup \left\{\left(\mathbf{y}_{j_t}, r(\mathbf{y}_{j_t})\right)\right\}, \quad & \left(\mathbf{y}_{j_t}, r(\mathbf{y}_{j_t})\right) \notin \mathcal{R}_{t}, \\
\mathcal{R}_{t}, \quad & \left(\mathbf{y}_{j_t}, r(\mathbf{y}_{j_t})\right) \in \mathcal{R}_{t}.\end{cases}
\end{align}


\noindent

\subsubsection{Objective Function}

Ideally, the traveler would design a policy that maximizes his total net gain based on the fact that any given decision $j_t$ is associated with a net gain determined by its underlying features, i.e., 

\begin{align} \label{eq:netgain}
g\left(S_t, j_t\right) = \begin{cases}
r(\mathbf{y}_{j_t})- c(\mathbf{x}_{e_t}), \quad & j_t \notin \mathcal{I}_t, \\
- c(\mathbf{x}_{e_t}), & j_t \in \mathcal{I}_t\setminus\{i_t\}, \\
0, &j_t = i_t.
\end{cases}
\end{align}

\noindent However, since no $c(\mathbf{x}_{e_t})$ and $r(\mathbf{y}_{j_t})$ are observed at the start of travel (i.e., $t=0)$,  the traveler's objective is to maximize his expected total net gain. His optimization problem is therefore to identify a policy that solves

\begin{align*}
\max_{\pi \in \Pi} \ \mathbbm{E} \left[ \sum_{t \in \mathcal{T}} g\left(S_t, J^\pi(S_t)\right) \right],
\end{align*}

\noindent where $J^\pi(S_t)$ is a decision function that returns $j_t$ based on the policy $\pi$, and $\Pi$ is the family of feasible policies under consideration. The expectation is calculated using the traveler's subjective beliefs as described in Section \ref{sec:TravelersPrior}.

\section{Problem Insights from a Clairvoyant Traveler}\label{sec:Insights}

The traveler's problem is complicated by the uncertainty about the node rewards and edge costs 
and by the combinatorial nature of the action space. 
Therefore, to inform policy design, we isolate the combinatorial challenge by considering a
clairvoyant traveler who has perfect knowledge about the graph, costs, and rewards.
This clairvoyant seeks to plan a walk that maximizes his net gain. 

The clairvoyant's problem is related to the longest-path problem \citep{karger1997approximating}.
This relationship can be seen by transforming $\mathcal{G}$ into a directed graph
and adding the node reward to each of its edges. Directed edges are necessary because, depending upon the direction in which an edge is traversed, 
the net gain may be distinct based on the different destination node rewards. 
The clairvoyant need not consider the features $\mathbf{x}_e$ and $\mathbf{y}_j$ since 
the rewards and costs are known. 
Therefore, by setting all node rewards to zero and the edge weights to $r_j - c_{(i,j)}$ 
where $r_j\ge 0$ is the known reward for node $j$ and $c_{(i,j)}\ge 0$ is the known traversal 
cost for directed edge $(i,j)$, the structural similarity of the clairvoyant's problem to the
longest-path problem is clear.  

However, unlike the longest-path problem, the clairvoyant's edge weights are not static. 
After node $j$ has been visited, the weights update to $- c_{(i,j)}$. 
Unfortunately, this additional structure does not reduce the computational complexity. 

The decision variant of the clairvoyant's problem asks whether there exists a walk yielding some pre-specified reward. 
Allowing this reward to equal $|\mathcal{N}|$, Property \ref{prop:NPHard} shows that, 
{\color{black}like the longest-path problem, finding the Hamiltonian path  reduces to the clairvoyant's problem \citep{karger1997approximating}.
The existence of such a Karp reduction \citep[e.g., see][]{zhang2022karp} implies} the clairvoyant's problem is intractable.\\

\begin{property} \label{prop:NPHard}
The clairvoyant's decision problem is NP-hard. 
\end{property}
\small
\noindent {\it Proof:} Take a Hamiltonian path problem on an undirected graph 
$\mathcal{H}=(\mathcal{N}_{\mathcal{H}},\mathcal{E}_{\mathcal{H}})$. 
Build a polynomial time reduction to the clairvoyant's problem as follows: 
add a fully connected node $i_0$ with $c_{(i_0,j)}=0, \ \forall j \in \mathcal{N}_{\mathcal{H}}$ 
and $r_{i_0} = 0$. 
Let $i_0$ be the clairvoyant's starting point, and set $c_{(i,j)}=1, \ \forall(i,j) \in 
\mathcal{E}_{\mathcal{H}}$ and $r_i = 2, \ \forall i \in \mathcal{N}_{\mathcal{H}}$. 
Call the new graph $\mathcal{G}=(\mathcal{N},\mathcal{E})$.

A Hamiltonian path solves the clairvoyant's problem:  Given a valid Hamiltonian path $\rho$ on $\mathcal{H}$, 
let him walk through $\mathcal{G}$, moving from $i_0$ to the endpoint of $\rho$. 
Since $|\mathcal{N}_H| =|\mathcal{N}| -1$,  this walk earns a reward of 
$2(|\mathcal{N}|- 1) - (|\mathcal{N}|-2)  = |\mathcal{N}|$, indicating a
solution to the clairvoyant's decision problem. 

The solution to the clairvoyant's problem is a Hamiltonian path: Assume his walk on 
$\mathcal{G}$ has reward $|\mathcal{N}|$. 
Such a walk cannot revisit a node, else its payoff is less than $|\mathcal{N}|$. 
Therefore, since traveling from $i_0$ to an unvisited node along $\rho$ yields a net gain of 2, 
and traveling to a unvisited node in $\rho$ from another along $\rho$ yields a net gain of 1, 
the clairvoyant's walk, excluding $i_0$, is a Hamiltonian path on $\mathcal{H}$. 
\hfill $\blacksquare$ \\

Property~\ref{prop:NPHard}'s computational challenge for the clairvoyant
applies to the traveler's problem in Section \ref{sec:ProbForm}. 
If the traveler is at $i_t$, he will 
seek an expected-utility-maximizing walk in the same manner as the clairvoyant. 
This strategy is common in sequential decision problems but it is 
NP-hard in the traveler's setting. 

Additional analysis of the clairvoyant's problem reveals that the network's payment structure
ensures some walks are never optimal. 
Properties \ref{prop:Circuit2x} and \ref{prop:DominatedCircuit} provide simple but useful 
results that reduce the clairvoyant's decision space. 
Moreover, these results also ensure that an optimal walk of finite length always exists. 
Properties \ref{prop:Circuit2x} and \ref{prop:DominatedCircuit} directly extend to the 
traveler's problem by replacing random edge costs and node rewards with their expectations. 

{\color{black}Let a circuit $\xi$ on $\mathcal{G}$ be a sequence of edges in which the first and last nodes 
are identical.
Nodes in a circuit may be repeated but edges may not.} 
Let $\mathcal{N}_\xi$ be the set containing distinct nodes visited in $\xi$. 
The following properties can be derived.\\

\begin{property} \label{prop:Circuit2x}
The clairvoyant's optimal walk does not repeat the same circuit. 
\end{property}
\small
\noindent {\it Proof:} Once a circuit is traversed, all its node rewards are collected. 
Repeating the circuit yields a negative net gain and returns the traveler to the same location.  \hfill $\blacksquare$ \\
\normalsize

\begin{property} \label{prop:DominatedCircuit}
If there exists distinct circuits $\xi_1$ and $\xi_2$, with the same end point and $\mathcal{N}_{\xi_1} \subset \mathcal{N}_{\xi_2}$, $\xi_1$ and $\xi_2$ cannot both be traversed on the clairvoyant's optimal walk. 
\end{property}

\small
\noindent {\it Proof:} Since 
\begin{align*}
\sum_{i \in \mathcal{N}_{\xi_2}} r_i- \sum_{(i,j) \in \xi_1 \cap \xi_2} c_{(i,j)} - \sum_{(i,j) \in \xi_2\setminus \xi_1} c_{(i,j)}  >&  \sum_{i \in \mathcal{N}_{\xi_2}} r_i - \ 2\sum_{(i,j) \in \xi_1 \cap \xi_2} c_{(i,j)} \\ & \quad - \sum_{(i,j) \in \xi_1\setminus \xi_2} c_{(i,j)} - \sum_{(i,j) \in \xi_2\setminus \xi_1} c_{(i,j)} ,
\end{align*}

\noindent it is more profitable to take only $\xi_2$ than both $\xi_1$ and $\xi_2$. And if 

\[
\sum_{i \in \mathcal{N}_{\xi_2} \setminus \mathcal{N}_{\xi_1} } r_i -\sum_{(i,j) \in \xi_2\ \setminus \xi_1} c_{(i,j)}  < -\sum_{(i,j) \in \xi_1\setminus \xi_2} c_{(i,j)}
\]

\noindent then it is more profitable to traverse $\xi_1$ than $\xi_2$. \hfill $\blacksquare$ \\
\normalsize

Properties \ref{prop:Circuit2x} and \ref{prop:DominatedCircuit} apply to any circuit, 
whether or not interior nodes are visited more than once. 
However, for circuits without repeated interior nodes, i.e.,  cycles, stronger results about the clairvoyant's 
optimal path can be derived. 
For large graphs, it is computationally burdensome to enumerate all cycles; 
however, they are characterized by a cycle basis. 
All cycles in $\mathcal{G}$ can be formed by symmetric differences of its basis cycles. 
Polynomial time algorithms exist to find such cycle bases under relatively mild assumptions 
\citep[i.e., see][]{horton1987polynomial}.  
But regardless of how it is computed, once a cycle basis is provided, the following results
apply to the clairvoyant's optimal path. \\

\begin{property} \label{prop:CycleBasis}
Any edge $(i,j)$ not present in a basis cycle of $\mathcal{G}$ is traversed at most twice in the clairvoyant's optimal walk. 
\end{property}
\small
\noindent {\it Proof: } If $(i,j)$ is not present in the cycle basis, it is not part of any 
cycle in $\mathcal{G}$. 
This implies $(i,j)$ is a bridge connecting two node sets that become disjoint if the 
edge were deleted. 
Call these node subsets $\mathcal{N}_1$ and $\mathcal{N}_2$ such that 
$\mathcal{N} = \mathcal{N}_1 \cup \mathcal{N}_2$. 

Without loss of generality, consider a finite walk $w_1$ with edge $(i,j)$ 
repeated $K$ times such that $i_0 \in \mathcal{N}_1$. 
Since $(i,j)$ is a bridge, we can write 
$w_1 = \left(w_{1,1}, (i,j), w_{1,2}, (i,j), ... , (i,j), w_{1,K+1} \right)$ where 
$w_{1,k}$ are walks contained within $w_1$. 
Note that $w_{1,k}$ only includes nodes in $\mathcal{N}_1$ when $k$ is odd, and 
$\mathcal{N}_2$ otherwise. 
Likewise, for odd $k>1$, the end points of $w_{1,k}$ are $i$, and for even $k<K+1$, 
the end points of $w_{1,k}$ are $j$.

If $K > 2$ is odd then $K+1$ is even, the walk $w_1$ terminates in $\mathcal{N}_2$, and 
we can construct a new walk $w_2 = \{w_{1,1}, w_{1,3}, ..., w_{1,K}, (i,j), w_{1,2}, 
w_{1,4},...,w_{1,K+1}\} $. 
This new walk visits the same nodes as $w_1$, but crosses $(i,j)$ only once, implying the total 
edge cost of $w_2$ is less than $w_1$. 
If $K> 2$ is even, we can similarly construct $w_2 = \{w_{1,1}, w_{1,3}, ..., w_{1,K-1}, (i,j),
\rho_{1,2}, \rho_{1,4},...,\rho_{1,K}, (i,j), \rho_{1,K+1}\}$ such that the cost of $w_2$ is 
less than that of $w_1$. 
Therefore, if $K>2$, another walk of lesser cost can always be constructed. \hfill $\blacksquare$ \\
\normalsize

Property \ref{prop:CycleBasis} implies that, if $\mathcal{G}$ contains a bridge, 
the clairvoyant can identify a large number of solutions that are dominated by the walk's 
structure alone. 
As the number of bridges in $\mathcal{G}$ increases, so too does the number of walks that can 
be discarded. 
A graph that maximally exhibits this property only includes edges that are bridges and, 
by definition, is acyclic. 
Property \ref{prop:CycleBasis} can be extended in such situations to generate further 
conditions on the clairvoyant's optimal walk. \\

\begin{property} \label{prop:TwoTimesDegree}
If $\mathcal{G}$ is acyclic, the number of times the clarivoyant visits a node on his 
optimal walk is bounded above by two times the node's degree.
\end{property}

\small
\noindent {\it Proof: } If $\mathcal{G}$ is acyclic, its cycle basis is empty and 
contains no edges. 
By Property \ref{prop:CycleBasis}, this implies every edge $(i,j) \in \mathcal{E}$ is 
traversed no more than twice. 
Therefore, any node $i \in N$ can be visited no more than two times its degree. \hfill $\blacksquare$ \\
\normalsize

\begin{property} \label{prop:AcyclicLength}
If $\mathcal{G}$ is acyclic, the clarivoyant's optimal walk does not exceed length 
$2|\mathcal{E}|$.
\end{property}

\small
\noindent {\it Proof: } A walk of greater length implies some $i \in N$ was visited 
more than two times its degree, thereby violating Property \ref{prop:TwoTimesDegree}.   
\hfill $\blacksquare$ \\
\normalsize

Property \ref{prop:AcyclicLength} provides an upper-bound on how many steps into the future 
the clairvoyant must plan in order to identify an optimal solution. 
But the results rely upon an acylic $\mathcal{G}$. 
Other properties enable upper bounds for general graphs, but with substantially more effort. 
The clairvoyant would need to account for all circuits associated with each node and 
determine which are dominated, and the effort to identify such bounds may exceed their utility.

Properties \ref{prop:TwoTimesDegree} and \ref{prop:AcyclicLength} imply that efficient 
algorithms may exist for the clairvoyant's problem when $\mathcal{G}$ is acyclic. 
Property \ref{prop:NPHard} ensures that, for general graphs, the problem is difficult to solve.
These results directly apply to the traveler's problem and to his corresponding policy design. 
On a general graph, optimal decisions are too computationally difficult to identify. 

\section{Solving the Traveler's Problem} \label{secAlgs}

Using the results from Section \ref{sec:Insights}, we examine four heuristic 
policies to solve the traveler's problem. 
These policies rely upon the traveler's belief at time $t$, which is calculated using 
Equations \eqref{eqPosterior} through \eqref{eqPostR}. 
At each new state $S_t$, the policy's associated decision function is used to determine 
the traveler's action based on his updated beliefs. 
Each policy is based on the similarity of the traveler's problem to the longest-path problem;
however, they differ in how they compute forecasts of a decision's expected future returns. 

\subsection{Myopic Baseline Policy}

The myopic policy (i.e., policy $M$) is the baseline used for comparative analysis. 
This policy centers on the immediate outcomes of a given decision $j_t$, disregarding the
potential long-term effects. 
Policy $M$ transitions from node $i_t$ to $j_t$ via the following decision function:
\begin{equation}
J^{M}(S_t) = \argmax_{j_t \in \mathcal{N}(i_t)} \ \mathbbm{E} \left[ g\left(S_t, j_t\right)
\right], \label{eqUCBPolicy}
\end{equation}
where $\mathcal{N}(i_t$) is the set of nodes adjacent to $i_t$. 
The expectation is taken using the posterior distributions over 
$c(\mathbf{x}_{e_t})|\mathcal{C}_{t}$ and $r(\mathbf{y}_{j_t})|\mathcal{R}_{t}$. 
Since both $r(\mathbf{y}_{j_t})|\mathcal{R}_{t}$ and $c(\mathbf{x}_{e_t})|\mathcal{C}_{t}$ 
are independent and normally distributed, it immediately follows that 
\begin{equation}\label{eq:Expnetgain}
\mathbbm{E}\left[g\left(S_t, j_t \right)\right] = \begin{cases}
\mu_{r|\mathcal{R}_t}(\mathbf{y}_{j_t})- \mu_{c|\mathcal{C}_t}(\mathbf{x}_{e_t}), 
\quad & j_t \notin \mathcal{I}_t, \\
- \mu_{c|\mathcal{C}_t}(\mathbf{x}_{e_t}), & j_t \in \mathcal{I}_t\setminus\{i_t\}, \\
0, &j_t = i_t.
\end{cases}
\end{equation}
The myopic policy is exploitative, focusing only on immediate expected rewards 
without considering variability. 
In the context of this research, the myopic policy may be suitable when dense network 
connections make long-term computations expensive, or when node rewards substantially 
exceed edge-traversal cost, thereby reducing the need for exploring future states and 
alternative actions.

\subsection{Upper Confidence Bound Policy}

The upper confidence bound (UCB) policy (i.e., policy $UCB$) is based on an 
exploration-exploitation trade-off. 
It modifies the traditional multi-arm bandit problem to suit our setting. 
The idea is to balance the trade-off between choosing actions that the agent knows yield 
good rewards (exploitation) and taking a risk by trying actions that the traveler has not 
explored much (exploration) to possibly find better solutions. 
The UCB algorithm deals with this trade-off by favoring actions that are both rewarding 
and less explored.  
Specifically, the $UCB$ policy transitions from node $i_t$ to $j_t$ via the following 
decision function:
\begin{align}
J^{UCB}(S_t| \lambda) = \argmax_{j_t \in \mathcal{N}(i_t)} \ \mathbbm{E} 
\left[ g\left(S_t, j_t\right) \right] + \lambda \mathbbm{V} \left[ g\left(S_t, j_t\right) \right], 
\label{eqMyopicPolicy}
\end{align}
where $\lambda\ge0$ is a tunable exploration parameter.  
As indicated in the previous section, $r(\mathbf{y}_{j_t})|\mathcal{R}_{t}$ and 
$c(\mathbf{x}_{e_t})|\mathcal{C}_{t}$ are independent and normally distributed.  
This allows us to generate the cases provided in Equation \eqref{eq:Expnetgain} as well 
as the following:
\begin{align} \label{eq:Expnetvar}
\mathbbm{V} \left[ g\left(S_t, j_t \right) \right] = \begin{cases}
\sigma^2_{r|\mathcal{R}_t}(\mathbf{y}_{j_t}) + \sigma^2_{c|\mathcal{C}_t}(\mathbf{x}_{e_t}), 
\quad & \left(\mathbf{x}_{e_t}, r(\mathbf{y}_{j_t})\right) \notin \mathcal{R}_{t},
\left(\mathbf{x}_{e_t}, c(\mathbf{x}_{e_t})\right) \notin \mathcal{C}_{t},\\
\sigma^2_{c|\mathcal{C}_t}(\mathbf{x}_{e_t}), \quad & \left(\mathbf{x}_{e_t},
r(\mathbf{y}_{j_t})\right) \in \mathcal{R}_{t},  \left(\mathbf{x}_{e_t}, 
c(\mathbf{x}_{e_t})\right) \notin \mathcal{C}_{t},\\
\sigma^2_{r|\mathcal{R}_t}(\mathbf{y}_{j_t}), \quad & \left(\mathbf{x}_{e_t}, 
r(\mathbf{y}_{j_t})\right) \notin \mathcal{R}_{t},  \left(\mathbf{x}_{e_t}, 
c(\mathbf{x}_{e_t})\right) \in \mathcal{C}_{t},\\
0, & \left(\mathbf{x}_{e_t}, r(\mathbf{y}_{j_t})\right) \in \mathcal{R}_{t},  
\left(\mathbf{x}_{e_t}, c(\mathbf{x}_{e_t})\right) \in \mathcal{C}_{t}.
\end{cases}
\end{align}

Different values of $\lambda$ cause different traveler choices. 
If $\lambda=0$, the traveler ignores the uncertainty in his estimate and moves to the node 
with the largest, immediate, expected net gain. 
It is equivalent to using the myopic policy. 
But if $\lambda$ is relatively large, it strongly encourages exploration, potentially 
forgoing immediate rewards but possibly leading to better long-term outcomes as it discovers 
higher-reward actions. 
The ideal $\lambda$ value depends on the specific environment and the problem at hand, 
balancing between immediate and future rewards.

\subsection{H-Path Policy} \label{secHPath}
The $H$-path policy (i.e., policy $HP$) is a rolling horizon policy, choosing an
action based on a preferred path of length $H$ from $i_t$. 
Using the information at $S_t$, policy $HP$ picks a preferred path and 
the traveler moves to the first node on that path. 
Upon reaching the new node, the traveler updates his priors and identifies a new preferred 
path of length $H$ and repeats the selection process. 


Similar to policy $UCB$, the $H$-path policy chooses actions by balancing the 
expected net gain against the reduction in uncertainty.  
It finds path $e^*_{1:H}$ from $i_t$ by solving 
\begin{align*} \label{gv}
\textbf{P1}:& \max_{\centering \myatop{e'_{1:H} \in \mathcal{P}_H(i_t)} {j'_k= d(e'_k), \ \forall k}} 
\  \mathbbm{E} \left[ \sum_{k=1}^H g\left(S'_k, j'_k\right) \right] + \alpha 
\left(\mathbbm{V}_G \left[ c\left(\mathbf{x}_{e'_{1:H}}\right) \big\vert \mathcal{C}_t \right] 
+ \mathbbm{V_G} \left[ r\left(\mathbf{y}_{j'_{1:H}}\right) \big\vert \mathcal{R}_t \right] 
\right), 
\end{align*}
where $S'_{1:H}$ is a sequence of synthetic states, $e'_{1:H}$ is a candidate path of 
length $H$ from $i_t$, $\mathcal{P}_H(i_t)$ is the set of all $H$-length paths from 
$i_t$,  $j'_ k= d(e'_k)$ such that $d(\cdot)$ provides the destination (tail) of an edge, 
$\alpha$ is a tunable exploration parameter, and
$\mathbbm{V}_G$ is the generalized variance (a scalar measure 
of multidimensional scatter). 
The smaller the value of $\mathbbm{V}_G$, the more association exists between the variables, 
so observed variables lead to greater reduction in uncertainty for the unobserved. {\color{black} For large $H$, the feasible region of Problem P1 will contain many paths; however, depending upon the structure of $\mathcal{G}$, Properties \ref{prop:Circuit2x} -- \ref{prop:AcyclicLength} help discard solutions based upon the already traversed walk.} 

Notably, the transition model for the synthetic future states differs slightly from that 
discussed in Section \ref{sec:ProbForm}; i.e., $S'_1 = S_t$, $S'_{k+1} = \langle j'_k, \mathcal{I}'_{k},
\mathcal{C}_t, \mathcal{R}_t  \rangle $ for $k<H$, and $\mathcal{I}'_{k+1}$ is updated similarly to 
equation \eqref{eq:I_t}. 
Therefore, only the traveler's position and history of visited nodes are updated. However, since revisiting nodes is prohibited on paths, if the decision space is properly constrained, one may forgo updating $\mathcal{I}_k$ as well.


By the linearity of expectation and the definition of generalized variance \citep{bagai1965distribution}, 
the expectation in Problem P1 can be simplified using Equation \eqref{eq:Expnetgain}
\begin{align*}
    \mathbbm{V}_G \left[ c\left(\mathbf{x}_{e'_{1:H}}\right) \big\vert \mathcal{C}_t \right] & = \det\left(\text{cov}_{c|\mathcal{C}_t}\left(\mathbf{x}_{e'_{1:H}}\right) \right), \\
    \mathbbm{V}_G \left[ r\left(\mathbf{x}_{j'_{1:H}}\right) \big\vert \mathcal{R}_t \right] & = \det\left(\text{cov}_{r|\mathcal{R}_t}\left(\mathbf{y}_{j'_{1:H}}\right) \right),
\end{align*}
where these covariance matrices are defined in Section \ref{sec:TravelersPrior}. 
Once Problem P1 is solved and the preferred path $e^*_{1:H}$ is identified, the $H$-path policy's decision
rule dictates the traveler move along $e^*_1$ to $d(e^*_1)$.  
Specifically, policy $HP$ transitions from node $i_t$ to $j_t$ via the following decision function: 
\begin{align}
J^{HP}(S_t| \alpha, H) = d(e^*_1).
\end{align}
Unfortunately, depending on the structure of $\mathcal{G}$ and the size of $H$, solving Problem P1 
can be computationally intractable. 
It is an NP-Hard problem. 
To identify $e^*_{1:H}$ at any state $S_t$, the traveler must address a regularized, single-source, 
longest-path problem. 
While feasible for acyclic $\mathcal{G}$ or small $H$ values, complex graphs or large $H$-values 
require alternative methods to approximate the optimal solution of Problem P1.

We use a neighborhood search heuristic to approximate a solution to Problem P1, thereby 
facilitating the application of the $H$-path policy for general problems. 
{\color{black}This heuristic employs a 2-Opt-like technique, adapting a method frequently used to solve traveling 
salesman problems \citep{hougardy2020approximation} whereby two edges on an existing walk are removed and the walk is reconnected with two new edges and their possible reordering.}
The algorithm for identifying this approximate solution, $\hat{e}_{1:H}$, is detailed in Algorithm \ref{algNS}. 

\begin{algorithm}[!htbp]
\caption{$H$-Path Neighborhood Search} 
\label{algNS}
\begin{algorithmic}[1]
\small
\State{\textbf{Input}: $\mathcal{G}$, $\mu_c$, $\Sigma_c$, $\mu_r$, $\Sigma_r$} 
\State{Generate initial feasible path $e'_{1:H}$}
\While{improvements can be made}
    \For{each pair of adjacent edges in $e'_{1:H}$}
        \State{remove the pair from $e'_{1:H}$}
        \State{identify the first substitution pair that improves objective function}
        \State{replace the removed pair with the identified substitution}
    \EndFor
    \State{update $e'_{1:H}$ with the modified path}
\EndWhile
\State{\textbf{Output}: $J^{HP}(S_t| \alpha, H) = d(e'_1)$}

\end{algorithmic}
\end{algorithm}
\normalsize

Beginning with a feasible path $e'_{1:H}$, Algorithm \ref{algNS} sequentially alters the path's visited nodes.
It navigates to neighboring solutions in $\mathcal{P}_H(i_t)$ by removing two adjacent edges from $e'_{1:H}$ 
and substituting them with the first identified pair that results in a new path with a superior objective
function value. 
The process continues iteratively with this new path, modifying all adjacent edges along the path until 
no further improvements can be found. 
Upon conclusion, the algorithm chooses an action tied to the first node in the walk linked with the 
optimized path $e'_{1:H}$.

\color{black}

\subsection{Speculating Clairvoyant Policy}
The speculating clairvoyant policy (i.e., policy $SC$) assumes that point estimates represent true 
values and constructs a policy mimicking the foresight of a clairvoyant.  
In this framework, when the traveler is at state $S_t$, the traveler's policy is constructed by 
using point estimates of the unknown parameters, and then solving the subsequent problem in a manner 
similar to a clairvoyant. 
The mean function estimates replace the unknowns in the graph, and, once the optimal clairvoyant 
solution is identified, the traveler progresses to the first node along this walk. 

The traveler attempts to solve the following optimization problem and, akin to the $H$-path 
policy, moves to the first node along the associated walk. 
The problem is
\begin{align*}
\textbf{P2}:& \max_{ \myatop{e'_{1:V} \in \mathcal{W}_V(i_t)} {j'_k= d(e'_k), \ \forall k}} \  
\mathbbm{E} \left[ \sum_{k=1}^{V} g\left(S'_k, j'_k\right) \right] 
\end{align*}
where $V$ is a sufficiently large number  (i.e., akin to the big-$M$ method in linear programming; cf.\
\citet{taha2013operations}), 
$\mathcal{W}_V(i_t)$ denotes the set of all $V$-length walks from $i_t$, and the possible state 
transitions are identical to the $H$-path policy. {\color{black} Note that, in general, $\lVert \mathcal{W}_V(i_t)\rVert$ can be large, but Properties \ref{prop:Circuit2x} -- \ref{prop:AcyclicLength} may be used to reduce its cardinality by discarding walks, depending upon the structure of $\mathcal{G}$ (e.g., those repeating the same circuit many times).} Once Problem P2 is solved, the associated decision rule is
\begin{align}
J^{SC}(S_t) = d(e^*_1),
\end{align}
such that $e^*_{1:V}$ is the optimal walk. 
Unfortunately, Property \ref{prop:NPHard} shows that solving Problem P2 is NP-hard, except for well-behaved graphs. 
Drawing on the parallels between the clairvoyant's problem and the longest path problem, 
we set forth a label-setting heuristic, inspired by the Bellman-Ford algorithm for the single-source, 
shortest-path problem, given in Algorithm \ref{algLS}.
\begin{algorithm}[!htbp]
\caption{Label-Setting Heuristic} 
\label{algLS}
\begin{algorithmic}
\State{\textbf{Input}:  $\mathcal{G}$, $S_t$, $\mu_c$, $\mu_r$} 
\State{Set $\mathcal{E}_D$ to the directed equivalent of $\mathcal{E}$}
\For{$k= 1, ..., \beta$}
\State{$\Omega_k(i) = \{\}, \ \forall i \in \mathcal{N}$} \Comment{No predecessors assigned}
\State{$z_k(i_t)=0$} 
\State{$z_k(i)=-\infty, \ \forall i \in \mathcal{N}\setminus {i_t} $}
\State{Randomly order $\mathcal{E}_D$ to create $\mathcal{E}'_D$}
\For{$n = 1, ..., N$}
\For{each $(i,j) \in \mathcal{E}'_D$}
\If{$j \notin \mathcal{I}_t$ \textbf{and} $j \notin \Omega(i)$}{ $\delta = \mu_{r|\mathcal{R}_t}(y_{j}) - \mu_{c|\mathcal{C}_t}(x_{(i,j)})$}
\Else{ $\delta = -\mu_{c|\mathcal{C}_t}(x_{(i,j)})$}
\EndIf
\If{$z_k(i) + \delta > z_k(j) $} \Comment{Associated walk yields a greater reward}
\State{$z_k(j) = z_k(i) + \delta$}
\State{$\Omega_k(j)= \Omega_k(i) \cup \{i\}$}
\EndIf
\EndFor
\EndFor
\EndFor
\State{$\{k^*, i^*\} = \argmax_{k,i} z_k(i)$}
\State{Set $e'_{1:V}$ according to node order of $\Omega_{k^*}(i^*)$ with $V= |\Omega_{k^*}(i^*)|$}
\State{\textbf{Output}: $J^{SC}(S_t) = d(e'_1)$}
\end{algorithmic}
\end{algorithm}

Algorithm \ref{algLS} uses a modified Bellman-Ford routine within an inner search loop that is repeated 
$\beta$ times.  
This inner search routine contrasts with the traditional Bellman-Ford algorithm via its labeling system 
and the direction of its inequalities. 
Here, labels encompass the entire sequence of visited predecessors during the walk. 
Classic assumptions, such as the prohibition against cycle traversal, no longer apply. 
Furthermore, while Bellman-Ford exclusively considers paths, Algorithm \ref{algLS} is able to 
handle more diverse walks. 
But, in so doing, the order in which edges are searched plays a more prominent role, potentially 
leading the algorithm to converge to a local optimum. 
Therefore, to manage these disparities, Algorithm \ref{algLS} incorporates a tuning parameter 
$\beta$ and edge order randomization. 
Namely, the algorithm uses a different random seed for every iteration $k$ to shuffle the 
order in which the edges are searched over the $\beta$ iterations. 
It induces stochasticity, facilitates exploration, and allows computational effort to be tuned.  
Upon concluding the $\beta$ iterations, the algorithm chooses an action tied to the first node in the 
walk linked with the maximum $z_k$-value.

\color{black}

\section{Computational Experimentation} \label{sec:CompResults}

This section {\color{black}empirically explores the Bayesian graph traversal problem and examines} the performance of the four previous policies.
Section \ref{sec:Illustrate} considers an example that shows the results of each policy and 
compares their performance to the optimal walk of a clairvoyant. {\color{black} Section \ref{secUAS} further examines policy performance on a more complicated instance based upon a public-safety case study to illustrate the practical implications of the Bayesian graph traversal problem. Thereafter,} Section \ref{sec:LargerTest} compares the \textcolor{black}{policies} to the myopic baseline on larger-scaled instances 
that cannot readily be solved, even by a clairvoyant. 
Random networks are generated based on the Erd\"os-R\'enyi model, and a full-factorial design over 
the parameters explores the effect of the network's structure on policy performance. 
All experiments and analyses are done on a dual Intel Xeon E5-2650v2 workstation with 128 GB of RAM
using MATLAB. In all experiments, $T=500$ so that the time horizon is not a limiting factor. 

\subsection{Illustrative Example} \label{sec:Illustrate}

\begin{figure}[htbp!]
    \centering
    \includegraphics[width=145mm]{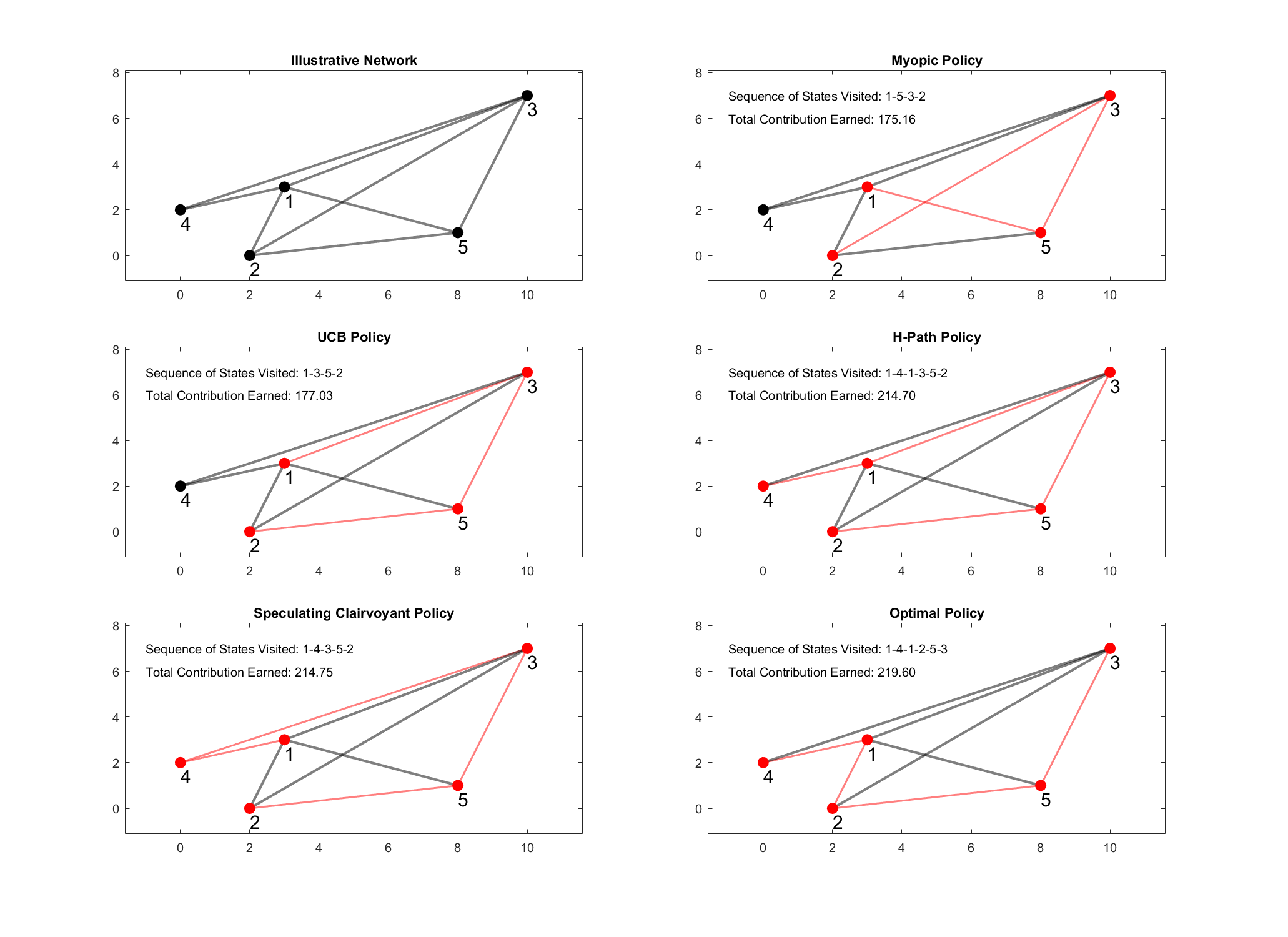}
    \caption{Illustrative Example - Policy Performance Comparison}
    \label{fig:ill}
\end{figure}

We begin by comparing the myopic policy to the optimal clairvoyant solution. 
The problem involves a densely connected network of five nodes.  
Each node $i$ is randomly situated within a two-dimensional Cartesian coordinate space of 
$(x_i, y_i) \in P$, where $P= \{(x, y): 0 \leq x, y \leq 10\}$. 
The network structure is shown in Figure \ref{fig:ill}. 
For legibility, a node's self-edges are not depicted, but their use is implied when the 
traveler ceases to move.

Here, the node characteristics (features) are the node's degree and the average degree of 
its neighbors. 
A node's degree is its number of edges.
The average neighbor degree, a metric employed in network structure analysis, 
reflects the connectivity of nodes linked to the given node. 
Edge features are the Cartesian coordinates of the nodes the edge connects.  
The node reward is derived from a second-order interaction model, with the first term as node degree, 
and the second as the average neighbor degree. 
The edge costs are a function of the Euclidean distance between linked nodes. 
Tables \ref{tab:node} and \ref{tab:edge} show the features and rewards/costs for the nodes and edges,
respectively. 

\begin{table}[htbp!]
\centering
\caption{Node Details}
\label{tab:node}
\begin{tabular}{ccccc}
\toprule
Label & Coordinates & Degree & Avg. Neighbor Degree & True Reward \\
\midrule
1 & (3,3) & 5 & 4.33 & $0^*$ \\ 
2 & (2,0) & 4 & 4.40 & 61.36 \\
3 & (10,7) & 5 & 4.33 & 74.78 \\
4 & (0,2) & 3 & 4.00 & 44.00 \\
5 & (8,1) & 4 & 4.40 & 61.36 \\
\bottomrule
\multicolumn{5}{l}{\tiny{*Without loss of generality, the reward associated with the starting node is zeroed out}}
\end{tabular}
\end{table}

\begin{table}[htbp!]
\centering
\caption{Edge Details}
\label{tab:edge}
\begin{tabular}{cr}
\toprule
Edges & True Cost \\
\midrule
$(1,2) \land (2,1)$ & 3.16 \\
$(1,3) \land (3,1)$ & 8.06 \\
$(1,4) \land (4,1)$ & 3.16 \\
$(1,5) \land (5,1)$ & 5.38 \\
$(2,3) \land (3,2)$ & 10.63 \\
$(2,5) \land (5,2)$ & 6.08 \\
$(3,4) \land (4,3)$ & 11.18 \\
$(3,5) \land (5,3)$ & 6.32 \\
\bottomrule
\end{tabular}
\end{table}

The traveler's beliefs about these unknown functions are defined in Section \ref{sec:TravelersPrior}.
Drawing from the Bayesian optimization literature \citep{frazier2018tutorial,garnett2023bayesian}, the prior mean functions are 
assumed to be constant. 
The mean functions for rewards and costs are the mean of all node rewards and edge-traversal costs.
From Tables~\ref{tab:node} and \ref{tab:edge}, the prior mean on the node rewards is 63.26, including the 
already-received reward from node one, and the prior mean on the edge costs is 6.75. 
So movement along the graph is generally profitable, and the myopic policy may perform relatively well. 

For the covariance functions for rewards and costs, we use a Radial Basis Function (RBF) kernel with the 
bandwidth parameter set to one \citep[i.e., see][ for more details]{buhmann2000radial, frazier2009knowledge}. 
This kernel implies that points near each other in the input space are highly correlated, 
capturing smooth transitions in our reward and cost structures across the network. 
Different mean and covariance functions can be explored, but the goal of this study is an initial analysis of 
the proposed policies under various algorithmic settings and problem characteristics. 
Parameter settings under consideration for policies $UCB$, $HP$, and $SC$  are shown in Table \ref{tab:illex}.
Algorithm \ref{algNS} is used as an optimization routine to solve the subproblems for policy $HP$.

Figure~\ref{fig:ill} and Table~\ref{tab:illex} display the results. 
Figure~\ref{fig:ill} shows the routes taken by each policy on the network as well as the amount
earned along the route. 
The optimal policy (i.e., the clairvoyant's route) has value 219.60. 
This value is as an upper bound on the traveler's net gain and is the reference point in Table~\ref{tab:illex}. 
Policies $M$ and $UCB$ are comparable, both selecting a route with reward about 80\% of the optimum.
Neither performed any back-tracking nor did they revisit a node; however, the tendency of policy $UCB$
to explore led it to find a somewhat better solution than the myopic policy. 
Policies $HP$ and $SC$ found comparable routes that were substantially better than the myopic policy, 
improving by approximately 23\%. 
Both routes are nearly optimal; their optimality gaps are each less than 3\%. 

\begin{table}[htbp!]
\centering
\caption{Policy Settings and Performance Results}
\label{tab:illex}
\resizebox{\textwidth}{!}{
\begin{tabular}{llcccc}
\toprule
 & Parameter & Sequence of  & Total Contribution & Improvement over   & Percent  \\
Policy ($\pi$) & Setting(s) & States Visited &  Earned &  Myopic Policy  & Optimal \\
\midrule
$M$ & - & 1-5-3-2 & 175.16 & - & 79.76\% \\
$UCB$ & $\lambda = 1$ & 1-3-5-2 & 177.03 & 1.07\% & 80.61\% \\
$HP$ & $\alpha = 1$; $H=3$ & 1-4-1-3-5-2 & 214.70 & 22.58\% & 97.77\% \\
$SC$ & $\beta = 1$ & 1-4-3-5-2 & 214.75 & 22.60\% & 97.79\% \\
Optimal ($\pi^*$) & - & 1-4-1-2-5-3 & 219.60 & 25.37\% & - \\
\bottomrule
\end{tabular}}
\end{table}

Note that, in order to maximize his value, the traveler must backtrack across edge $(1,4)$.
However, backtracking behavior will never result from policies $M$ or $UCB$; no reward is given for 
revisiting a node and no knowledge is obtained by recrossing an edge. 
Furthermore, revisiting a node via an untraversed edge is impossible in policy $M$ and generally discouraged in policy $UCB$. 
The former is true because no reward is gained by revisiting a node, and the latter holds so long as 
little information is gained from crossing the untraversed edge. 
No such restrictions are present in policies $HP$ and $SC$. 




{\color{black}
\subsection{Case Study: UAS Operations for Public Safety} \label{secUAS}
The results from the previous section lead us to explore how our policies might perform in a more realistic setting. Therefore, we perform an additional empirical exploration in a public-safety context. 
Specifically, we show how Bayesian graph traversal relates to the use of unmanned aerial systems (UASs) 
for first responders and how the policies can be used for that application. 

Small UASs are increasingly viewed as force multipliers in public-safety operations. 
Their use has been advocated for law enforcement by the \citet{IACP} to enhance situational awareness,
investigate crime scenes, and respond quickly to disturbances while police officers are en route 
(e.g., via equipped two-communication).  
Similar capabilities have been promoted by the \citet{IAFC} for structural and wildland firefighting, among other disaster response operations. 
This first-responder demand has grown to such a degree that numerous companies (e.g., Skydio, UVT, Brinc) 
have emerged to satisfy it, and the \citet{NUSTL2024}, a subordinate unit of the US Department of Homeland Security, has published implementation recommendations. 
Notably, in any public-safety application, UASs must orient and route themselves within an uncertain environment, implying a Bayesian graph traversal problem. 

To illustrate this connection, consider a police department in a standoff against criminals holding 
hostages on a compound. 
The compound contains three buildings and, although the police are uncertain of the configurations of the 
rooms, they have the blueprints of the structures. 
Hostage negotiators have been unsuccessful, so a SWAT team kinetic rescue operation is needed. 
The police have access to a drone with an on-board glass breaker (e.g., akin to the Brinc LEMUR) 
that is also equipped with other sensors (e.g., visible-light and thermal cameras). 
The police will use the drone to obtain information about the compound and the locations of the criminals 
and hostages, while minimizing the risk of detection risk. 
The police want the element of surprise and, if the criminals notice the drone's presence, 
their behavior may become erratic, threatening the lives of the hostages and compromising the 
SWAT team's objectives. 

\begin{figure}[htbp!]
    \centering
    \includegraphics[width=115mm]{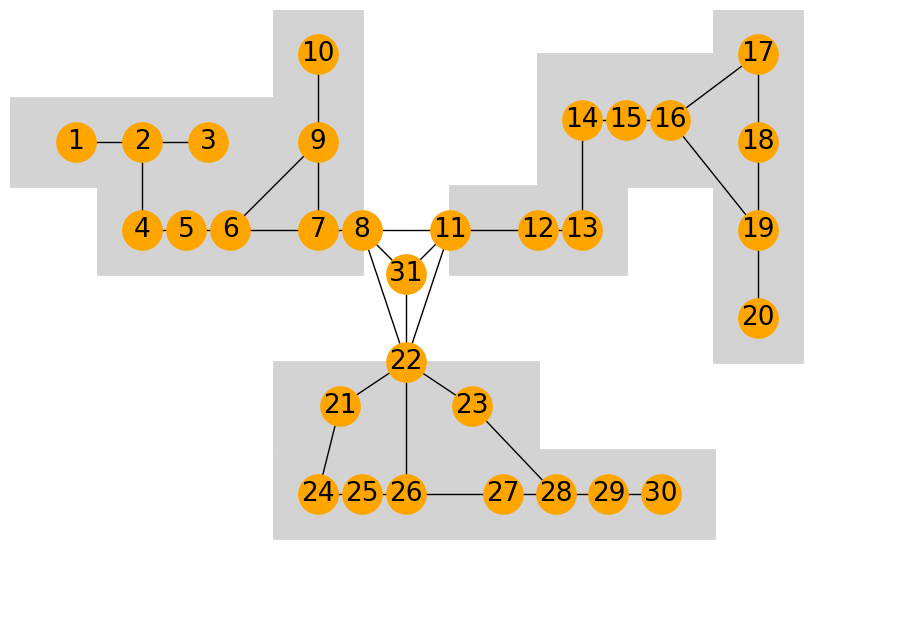}
    \caption{Graph Structure of Notional Compound Encountered by UAS Supporting Law-Enforcement Operations}
    \label{fig:compound}
\end{figure}

Figure \ref{fig:compound} shows the information from the compound's blueprints. 
Nodes represent locations of interest, and edges represent possible connections between them. 
There are multiple doors and windows into each building; however, the police have identified three
windows that the UAS could break with minimal noise (i.e., Nodes 8, 11, and 22). 
Prior to breaching a window, the police will create an auditory distraction so that traversing window 
edges is no more risky than traversing any other. 

As in Section \ref{sec:Illustrate}, node features consist of the node degree and the average neighbor degree. 
Thus, the police assume that rooms with heavier traffic contain more information than those less visited, 
and that the criminals' staging locations are based upon the compound design. 
Edge features are related to the Euclidean distance between the edge's tail and head with the 
assumption that longer transit times incur increased detection risk. 
The utility for rewards and costs is again assumed to be additive, with values assigned using the 
formulas in Section \ref{sec:CompResults}. 
Node rewards are formed using a second-order interaction model of the node features, and edge costs 
are a function of the Euclidean distance between nodes. 
The UAS's prior mean functions are set equal to their respective grand means, and RBF kernels with a 
bandwidth of 1 are the covariance functions. 
The UAS attempts to gain as much utility as possible and, rather than incur additional detection risk via
exploration, will bed down in a final location to await the SWAT intervention. 
Assuming the UAS begins at Node 31, we explore the performance of each policy in this 
situation.

Building off results from Section \ref{sec:Illustrate}, the same baseline policy parameter settings are 
used by the UAS. 
Table \ref{tab:baseline_UAS} shows the performance of each baseline. 
Whereas the size and geometry of the graph provided in Figure \ref{fig:compound} complicate the 
identification of an optimal clairvoyant walk, preventing the identification of a policy's optimality gap, 
the myopic policy may still be used as a benchmark. 
The baseline UCB policy provides no improvement to this benchmark; it identifies the same walk as the myopic 
policy. 
But policies $HP$ and $SC$ improve over policy $M$, by differing amounts. 
Policy $HP$ showed relatively minor improvement, but the improvement of policy $SC$ was more significant. 
Such results exhibit clear qualitative behaviors in each policy's walk as well. 
Policy $M$ and $UCB$ never enter the south building, but did extensive exploration of the northwest building.
Policy $HP$ visited all three buildings, but only peered inside without substantive exploration. 
Policy $SC$ visited all three buildings and conducted extensive surveillance of the south building, 
a beneficial choice given its relatively high density of larger rewards. 
All baseline policies neglected the northeast building where rewards are more sparse. 

\begin{table}[htbp!]
\centering
\caption{Performance of Baseline Policies in UAS Case Study}
\label{tab:baseline_UAS}
\resizebox{\textwidth}{!}{
\begin{tabular}{llcccc}
\toprule
 & Parameter & Sequence of  & Total Contribution & Improvement over  \\
Policy ($\pi$) & Setting(s) & States Visited &  Earned &  Myopic Policy   \\
\midrule
$M$ & - & 31-11-8-7-6-9-10 & 126.66 & -  \\
$UCB$ & $\lambda = 1$ & 31-11-8-7-6-9-10 & 126.66 & 0\%   \\
$HP$ & $\alpha = 1$; $H=3$ & 31-11-22-8 & 129.15 & 1.96\%  \\
$SC$ & $\beta = 1$ & 31-11-8-22-26-27-28-29-30 & 173.29 & 36.81\% \\
\bottomrule
\end{tabular}
}
\end{table}

Note the contrasting performance of policy $HP$ in Tables and \ref{tab:illex} and \ref{tab:baseline_UAS}.
It shows that a policy's success can vary and the need for hyperparameter tuning. 
The baseline policy $HP$ substantially improved over policy $M$ in Section \ref{sec:Illustrate}, 
but provided no improvement when routing the UAS. 
Therefore, to explore the effect of hyperparameter tuning on policy performance, we consider the effects of parameterizing policy $UCB$ with $\lambda \in \{1, 10\}$, policy $HP$ with $(\alpha, H) \in \{0,1,10 \} \times \{3,4,5\}$, and policy $SC$ with $\beta \in \{1,10,100\}$. 
The performance of these policies, along with their corresponding walks, are provided in Table \ref{tab:UAS_altpolicies}.


\begin{table}[htbp!]
\centering
\caption{Performance of Alternatively Parameterized Policies in UAS Case Study}
\label{tab:UAS_altpolicies}
\resizebox{\textwidth}{!}{
\begin{tabular}{llcccc}
\toprule
 & Parameter & Sequence of  & Total Contribution & Improvement over  \\
Policy ($\pi$) & Setting(s) & Nodes Visited &  Earned &  Myopic Policy   \\
\midrule
$M$ & - & 31-11-8-7-6-9-10 & 126.66 & -  \\
$UCB^*$ & $\lambda = 1$ & 31-11-8-7-6-9-10 & 126.66 & 0\%   \\
$UCB^{**}$ & $\lambda = 10$  & 31-11-8-22-26-27-28-23 & 161.73 & 27.69\% \\
$HP^*$ & $\alpha = 0$; $H=3$  & 31-11-22-8& 129.15 & 1.96\% \\
$HP^{**}$ & $\alpha = 0$; $H=4$  &31-11-31-8-22 & 135.83 & 7.24\% \\
$HP$ & $\alpha = 0$; $H=5$  &  31-11-22-11-12 & 42.40 & -66.53\%  \\
$HP$ & $\alpha = 1$; $H=3$ & 31-11-22-8 & 129.15 & 1.96\%  \\
$HP$ & $\alpha = 1$; $H=4$  & 31-11-31-8-22 & 135.83 & 7.24\%\\
$HP$ & $\alpha = 1$; $H=5$ &  31-11-22-11-12 & 42.40 & -66.53\%   \\
$HP$ & $\alpha = 10$; $H=3$  & 31-11-22-8-11 &  89.15&  -37.51\%\\
$HP$ & $\alpha = 10$; $H=4$   & 31-11-31-8-22 & 135.83 & 7.24\% \\
$HP$ & $\alpha = 10$; $H=5$   & 31 11 22 11 12 & 42.40 & -66.53\%\\
$SC^*$ & $\beta = 1$ & 31-11-8-22-26-27-28-29-30 & 173.29 & 36.81\% \\
$SC$ & $\beta = 10$ & 31-22-23-28-27-26-22-11-8-7 & 152.60 & 20.48\% \\
$SC^{**}$ & $\beta = 100$ & 31-22-23-28-27-26-25-24-21-22-11-8-7 & 207.88 & 64.12\% \\
\bottomrule
\multicolumn{5}{l}{\tiny{*Baseline parameter settings for each policy}} \\
\multicolumn{5}{l}{\tiny{**Best performing parameter settings for each policy}} \\
\end{tabular}
}
\end{table}



These results show that the performances of policies $UCB$, $HP$, and $SC$ were improved by exploring
alternative hyperparameter options. 
Improvement over policy $M$ was largest for policies $UCB$ and $SC$ with gains of approximately 27\% and 28\%,
respectively. 
Policy $HP$ improved as well. 
The substantive gains made by the tuned policy $UCB$ result from its exploration of the south building. 
The first seven nodes in its walk coincide with those of the baseline policy $HP$. 
This increased exploration makes sense given its larger value of $\lambda$. 
Alternatively, the tuned policy $HP$ chose a walk that more thoroughly explored both the south and 
northwest buildings. 
This behavior was driven by a larger $\beta$; which makes sense since increased $\beta$-values favor a 
longer search for each subproblem. 
However, based on the randomness of Algorithm \ref{algLS}, a larger $\beta$ does not guarantee 
improved performance---note the performance of policy $SC$ with $\beta=10$ in Table \ref{tab:UAS_altpolicies}.

The behavior of policy $HP$ is more nuanced. 
Multiple parameter choices yield walks of similar quality. 
Across every level of $\alpha$, the policy with $H=4$ performed best and identified the same walk. 
Similarly, whenever $H=5$, policy $HP$ substantively underperformed policy $M$. 
Thus, for this example, there is little benefit from varying the exploration parameter, 
$\alpha$; and by increasing $H$ to forecast too far ahead, the traveler's performance was degraded. 
This outcome raises questions regarding how the specifics of a problem instance affects policy performance, 
a topic we explore in the next section.


}

\subsection{Policy Performance on Random Networks} \label{sec:LargerTest}

This section studies different network structures to determine how they affect the policies' decisions. 
We generate many random networks and evaluate each policy over them. 
For each network, we use the same features for the node rewards and edges, and the same true reward and cost 
functions, as in Section \ref{sec:Illustrate}. 
{\color{black} The nodes for each instance are randomly placed on 
coordinates of 
$P$, the prior mean functions are defined as their respective grand means, and RBF kernels with  
bandwidth one are used as covariance functions. 
True node rewards and edge costs are also calculated as in Section \ref{sec:Illustrate} based upon the random 
graph structure and node locations.}

We generate random networks from the Erd\"os–R\'enyi model. 
Erd\"os–R\'enyi models have two parameters: the number of nodes $|\mathcal{N}|$ and the edge creation 
probability $p$ \citep{newman2018networks}. 
For a range of $|\mathcal{N}|$ and $p$ values, we generate networks, discarding graphs that are not
connected.  
We also ensure that each node has a self-loop, so that the exploration can terminate.
Table~\ref{tab:faclev} shows values $|\mathcal{N}|$ and $p$ that we explored.
{\color{black}Thirty connected graphs for each $|\mathcal{N}|$ and $p$ combination were generated and analyzed.}

Across each of the Erd\"os–R\'enyi settings, we examine the performance of the policy parameter 
settings explored in Section \ref{secUAS}.
Table~\ref{tab:faclev} shows the factor levels used for each of policy $UCB$, $HP$, and $SC$. 
These levels provide distinct incentives for each policy. 
For example, by increasing $H$ in policy $HP$, the traveler has additional forethought; 
increasing $\lambda$ in policy $UCB$ means the traveler explores more. 
{\color{black}Our goal is to identify how well a policy parameterization generalizes across similar
problems.}

\begin{table}[htbp!]
\centering
\caption{Parameter Values}
\label{tab:faclev}
\begin{tabular}{cl}
\toprule
\multicolumn{1}{c}{Parameter} & \multicolumn{1}{l}{Settings} \\
\midrule
$|\mathcal{N}|$               & \{20, 50, 80\}       \\
$p$               & \{0.2, 0.5, 0.8\}    \\ \cdashline{1-2} 
$\lambda$         & $\{0^*,1,10\}$       \\  
$H$               & \{3, 4, 5\}       \\ 
$\alpha$         & \{0, 1, 10\}       \\   
$\beta$         & \{1, 10, 100\}       \\   
\bottomrule
\multicolumn{2}{l}{\tiny{*Represents myopic policy}}
\end{tabular}
\end{table}


 
{\color{black}Table~\ref{expres} summarizes our experiment, indicating the best-performing policy parameters 
as well as statistical summaries of their performance over 30 draws from each Erd\"os–R\'enyi setting. 
Columns one and two specify the nine Erd\"os–R\'enyi parameter combinations.} Columns three to six identify the best settings for the $UCB$, $HP$, and $SC$ policies. 
Columns seven through nine report the 95\% confidence intervals on the average improvement over the 
myopic policy, giving us a metric for algorithmic effectiveness. 
Policy $M$ is used as a baseline instead of the clairvoyant's solution, which becomes intractable as the 
graph grows in complexity.

\begin{table}[htbp!]
\caption{Experimental Design Results}
\centering
\label{expres}
\begin{tabular}{cclrrrllrrr}
\toprule
\multicolumn{2}{c}{Erd\"os–R\'enyi} && \multicolumn{4}{c}{Best Algorithmic} && \multicolumn{3}{c}{Improvement Over}  \\ 
\multicolumn{2}{c}{Parameter Settings} && \multicolumn{4}{c}{Parameter Settings} && \multicolumn{3}{c}{Myopic Policy (\%)}  \\ 
\cline{1-2}	\cline{4-7} \cline{9-11} 
$|\mathcal{N}|$ & $p$ && $\lambda$ & $H$ & $\alpha$ & $\beta$ && \multicolumn{1}{c}{$UCB$} & \multicolumn{1}{c}{$HP$} & \multicolumn{1}{c}{$SC$} \\ \midrule
20	&	0.2	&&	-10	&	4	&	0	&	1	&&	$20.31\pm0.32$	&	$16.34\pm0.12$	&	$33.08\pm0.13$			\\
20	&	0.5	&&	-1	&	5	&	-10	&	100	&&	$1.54\pm0.14$	&	$23.51\pm0.21$	&	$20.81\pm0.26$			\\
20	&	0.8	&&	-10	&	3	&	-1	&	1	&&	$13.59\pm0.25$	&	$31.29\pm0.28$	&	$70.20\pm0.41$			\\
50	&	0.2	&&	-10	&	5	&	-10	&	100	&&	$16.72\pm0.47$	&	$40.53\pm0.52$	&	$29.38\pm1.44$			\\
50	&	0.5	&&	-10	&	3	&	-1	&	10	&&	$3.43\pm1.11$	&	$27.37\pm0.98$	&	$27.58\pm1.38$		\\
50	&	0.8	&&	-1	&	4	&	-10	&	100	&&	$-17.36\pm0.92$	&	$27.16\pm0.66$	&	$38.61\pm1.28$		\\
80	&	0.2	&&	-1	&	5	&	-10	&	10	&&	$27.28\pm1.33$	&	$53.38\pm0.93$	&	$22.93\pm2.06$	\\
80	&	0.5	&&	-1	&	5	&	-1	&	100	&&	$-3.47\pm1.18$	&	$26.63\pm1.57$	&	$32.52\pm2.01$	\\
80	&	0.8	&&	-10	&	5	&	-10	&	10	&&	$-5.29\pm0.68$	&	$19.43\pm1.73$	&	$99.56\pm2.48$\\ \bottomrule			
\end{tabular}
\end{table}

Table~\ref{expres} shows that policy $UCB$ does not always improve over policy $M$. 
The increased exploration generally pays off for simpler networks but not for complex ones.
Policies $HP$ and $SC$ consistently outperform the myopic policy for all nine problem cases, 
implying these are superior choices for the types of networks we study. 
As in Section~\ref{sec:Illustrate}, the improvement is often dramatic. 
Policy $HP$'s minimum average improvement was 16\% and its maximum was 54\%. 
Policy $SC$'s average improvement ranged from 20\% to nearly 100\%. 

Interpreting the relationship between network size and sparsity on policy success is difficult.
For $|\mathcal{N}|=20$, policy $HP$'s expected performance over policy $M$ increases as the graphs 
become more dense, whereas for $|\mathcal{N}|=80$, the opposite occurs. 
Policy $SC$ attains its best performances for the densest graphs examined in our experiments, 
with average improvement over $M$ most pronounced for $|\mathcal{N}|=20,80$, likely due to its ability to 
consider cycles. 
Finally, in contrast to policy $UCB$, policy $HP$ tends to benefit from exploration; in all but one 
Erd\"os–R\'enyi setting, $\alpha>0$ was preferred. 
A larger $\beta$ often benefited policy $SC$, but the randomness of its search meant that, 
in some settings, smaller $\beta$-values found better solutions. 

Although policy $HP$ and $SC$ are globally preferable to policy $UCB$, no single policy dominates.
The most effective policy depends upon the specific Erd\"os–R\'enyi parameters. 
The improvements over the myopic policy reinforce this finding, and the confidence intervals
quantify the variability in the outcomes. 
Moreover, there are no universally best policy parameter settings for all situations. 
Instead, parameter adjustments are necessary to optimize each policy, dependent on the unique 
conditions of each problem type.  
Nonetheless, significant improvement over policy $M$ is achievable given the correct implementation 
of either policy $HP$ or policy $SC$. 
This insight highlights the importance of an informed, context-sensitive approach to policy selection for each
application. 
It also suggests that the increased complexity in the subproblems generated by policies $HP$ and $SC$ 
can yield substantial dividends, justifying the increased computational effort.


\section{Conclusion} \label{sec:Conclusion}

We study movement of a traveler along a graph in which a random reward is gained upon the first visit to
a node and a random cost is incurred for each edge traversal. 
The traveler knows the graph's adjacency matrix and his starting position.
The Bayesian traveler has Gaussian process priors about the rewards and costs, leading to
a sequential decision-making problem that applies to many other situations than just
graph traversal (e.g., changing jobs, selecting college courses, investment in security,
convoy routing in hostile territory, portfolio management). 

The problem is NP-hard, and so one needs heuristic policies to find a solution.
We compared four policies that balance exploration and exploitation in different ways to heuristically 
optimize expected utility.
The performance of these policies was examined on a small network and on random (connected) graphs
from an Erd\"os–R\'enyi model whose parameters were varied in an experimental design.  


Additional theoretical extensions exist. 
For example, one may consider how the accuracy of the traveler's beliefs interacts with the policy type 
and network structure. 
It may be the case that some policies perform better when the traveler's priors are initially 
accurate (inaccurate) or when the network is dense (sparse). 
{\color{black} Such issues also drive extensions of this work that consider value-function approximation 
policies. 
The efficacy of pre-training such policies is affected by the accuracy of the traveler's beliefs, 
and future work may consider when such beliefs are accurate enough to warrant the additional 
computational burden. 
Moreover, in future studies, one could model the costs and rewards jointly, modify the formulation such to
consider noisy costs and rewards, or propose an adversarial framework in which
an opponent can perturb edge costs or node rewards. 
Adaptations of policies provided in other work \citep[e.g., see][]{brown2013optimal, thul2023information} or
applications to relevant scenarios in the literature \citep[e.g., see][]{suh2017fast,fossum2021learning, mccammon2021topological} may also be possible given their relationship to the Bayesian graph traversal problem.}
Finally, given that none of our policies dominates the others, future research could study alternative policies 
which balance solution quality against computational effort, or identify 
the network characteristics that should guide policy choice.

\small
\section*{Acknowledgments}
This work was supported by the Office of Naval Research Grant 6000012277 GRA as well as by Air Force Office of Scientific Research grant 21RT0867. 

\section*{Disclaimer}
The views expressed in this article are those of the authors and do not reflect the official policy or position of the United States Air Force, United States Department of Defense, or United States Government.

\normalsize


\bibliography{ref}

\begin{thebibliography}{41}
\providecommand{\natexlab}[1]{#1}
\providecommand{\url}[1]{\texttt{#1}}
\expandafter\ifx\csname urlstyle\endcsname\relax
  \providecommand{\doi}[1]{doi: #1}\else
  \providecommand{\doi}{doi: \begingroup \urlstyle{rm}\Url}\fi

\bibitem[Aksakalli et~al.(2016)Aksakalli, Sahin, and Ari]{aksakalli2016based}
Vural Aksakalli, O~Furkan Sahin, and Ibrahim Ari.
\newblock {An AO* Based Exact Algorithm for the Canadian Traveler Problem}.
\newblock \emph{INFORMS Journal on Computing}, 28\penalty0 (1):\penalty0
  96--111, 2016.

\bibitem[Al-Kanj et~al.(2023)Al-Kanj, Powell, and
  Bouzaiene-Ayari]{al2023information}
Lina Al-Kanj, Warren~B Powell, and Belgacem Bouzaiene-Ayari.
\newblock The information-collecting vehicle routing problem: Stochastic
  optimization for emergency storm response.
\newblock \emph{arXiv preprint arXiv:2301.06497}, 2023.

\bibitem[Bagai(1965)]{bagai1965distribution}
OP~Bagai.
\newblock {The Distribution of the Generalized Variance}.
\newblock \emph{The Annals of Mathematical Statistics}, 36\penalty0
  (1):\penalty0 120--130, 1965.

\bibitem[Bent and Van~Hentenryck(2004)]{bent2004scenario}
Russell~W Bent and Pascal Van~Hentenryck.
\newblock Scenario-based planning for partially dynamic vehicle routing with
  stochastic customers.
\newblock \emph{Operations Research}, 52\penalty0 (6):\penalty0 977--987, 2004.

\bibitem[Borrero et~al.(2016)Borrero, Prokopyev, and
  Saur{\'e}]{borrero2016sequential}
Juan~S. Borrero, Oleg~A. Prokopyev, and Denis Saur{\'e}.
\newblock {Sequential Shortest Path Interdiction with Incomplete Information}.
\newblock \emph{Decision Analysis}, 13\penalty0 (1):\penalty0 68--98, 2016.

\bibitem[Brown and Smith(2013)]{brown2013optimal}
David~B Brown and James~E Smith.
\newblock Optimal sequential exploration: Bandits, clairvoyants, and wildcats.
\newblock \emph{Operations research}, 61\penalty0 (3):\penalty0 644--665, 2013.

\bibitem[Buhmann(2000)]{buhmann2000radial}
Martin~Dietrich Buhmann.
\newblock Radial basis functions.
\newblock \emph{Acta Numerica}, 9:\penalty0 1--38, 2000.

\bibitem[Cohen et~al.(2021)Cohen, Efroni, Mansour, and
  Rosenberg]{cohen2021minimax}
Alon Cohen, Yonathan Efroni, Yishay Mansour, and Aviv Rosenberg.
\newblock {Minimax Regret for Stochastic Shortest Path}.
\newblock In M.~Ranzato, A.~Beygelzimer, Y.~Dauphin, P.~S. Liang, and
  J.~Wortman Vaughan, editors, \emph{Advances in Neural Information Processing
  Systems}, volume~34, pages 28350--28361. Curran Associates, Inc., 2021.

\bibitem[Fossum et~al.(2021)Fossum, Travelletti, Eidsvik, Ginsbourger, and
  Rajan]{fossum2021learning}
Trygve~Olav Fossum, C{\'e}dric Travelletti, Jo~Eidsvik, David Ginsbourger, and
  Kanna Rajan.
\newblock Learning excursion sets of vector-valued gaussian random fields for
  autonomous ocean sampling.
\newblock \emph{The annals of applied statistics}, 15\penalty0 (2):\penalty0
  597--618, 2021.

\bibitem[Frazier et~al.(2009)Frazier, Powell, and
  Dayanik]{frazier2009knowledge}
Peter Frazier, Warren Powell, and Savas Dayanik.
\newblock {The Knowledge-Gradient Policy for Correlated Normal Beliefs}.
\newblock \emph{INFORMS Journal on Computing}, 21\penalty0 (4):\penalty0
  599--613, 2009.

\bibitem[Frazier(2018{\natexlab{a}})]{frazier2018bayesian}
Peter~I. Frazier.
\newblock {Bayesian Optimization}.
\newblock In \emph{INFORMS TutORials in Operations Research: Recent Advances in
  Optimization and Modeling of Contemporary Problems}, pages 255--278. INFORMS,
  2018{\natexlab{a}}.

\bibitem[Frazier(2018{\natexlab{b}})]{frazier2018tutorial}
Peter~I. Frazier.
\newblock {A Tutorial on Bayesian Optimization}.
\newblock {arXiv preprint arXiv:1807.02811}, 2018{\natexlab{b}}.

\bibitem[Gardner et~al.(2014)Gardner, Kusner, Xu, Weinberger, and
  Cunningham]{gardner2014bayesian}
Jacob~R Gardner, Matt~J Kusner, Zhixiang~Eddie Xu, Kilian~Q Weinberger, and
  John~P Cunningham.
\newblock Bayesian optimization with inequality constraints.
\newblock In \emph{ICML}, volume 2014, pages 937--945, 2014.

\bibitem[Garnett(2023)]{garnett2023bayesian}
Roman Garnett.
\newblock \emph{{Bayesian Optimization}}.
\newblock Cambridge University Press, Cambridge, UK, 2023.

\bibitem[Han et~al.(2014)Han, Lee, and Park]{han2014robust}
Jinil Han, Chungmok Lee, and Sungsoo Park.
\newblock {A Robust Scenario Approach for the Vehicle Routing Problem with
  Uncertain Travel Times}.
\newblock \emph{Transportation Science}, 48\penalty0 (3):\penalty0 373--390,
  2014.

\bibitem[Horton(1987)]{horton1987polynomial}
Joseph~Douglas Horton.
\newblock {A Polynomial-Time Algorithm to Find the Shortest Cycle Basis of a
  Graph}.
\newblock \emph{SIAM Journal on Computing}, 16\penalty0 (2):\penalty0 358--366,
  1987.

\bibitem[Hougardy et~al.(2020)Hougardy, Zaiser, and
  Zhong]{hougardy2020approximation}
Stefan Hougardy, Fabian Zaiser, and Xianghui Zhong.
\newblock {The approximation ratio of the 2-Opt Heuristic for the metric
  Traveling Salesman Problem}.
\newblock \emph{Operations Research Letters}, 48\penalty0 (4):\penalty0
  401--404, 2020.

\bibitem[{International Association of Chiefs of Police}(2020)]{IACP}
{International Association of Chiefs of Police}.
\newblock {Small Unmanned Aircraft Systems}.
\newblock
  \url{https://www.theiacp.org/sites/default/files/2020-06/Unmanned%20Aircraft%20FULL%20-%2006222020.pdf},
  2020.

\bibitem[{International Association of Fire Chiefs}(2024)]{IAFC}
{International Association of Fire Chiefs}.
\newblock {UAS Tactics}.
\newblock
  \url{https://www.iafc.org/topics-and-tools/communications-technology/uas-toolkit/uas-resource/uas-tactics},
  2024.

\bibitem[Karger et~al.(1997)Karger, Motwani, and
  Ramkumar]{karger1997approximating}
David Karger, Rajeev Motwani, and Gurumurthy~DS Ramkumar.
\newblock On approximating the longest path in a graph.
\newblock \emph{Algorithmica}, 18\penalty0 (1):\penalty0 82--98, 1997.

\bibitem[Lagos et~al.(2024)Lagos, Auad, and Lagos]{lagos2024online}
Tom{\'a}s Lagos, Ram{\'o}n Auad, and Felipe Lagos.
\newblock The online shortest path problem: Learning travel times using a
  multiarmed bandit framework.
\newblock \emph{Transportation Science}, 2024.

\bibitem[Lim et~al.(2006)Lim, Bearden, and Smith]{lim2006sequential}
Churlzu Lim, J~Neil Bearden, and J~Cole Smith.
\newblock {Sequential Search with Multiattribute Options}.
\newblock \emph{Decision Analysis}, 3\penalty0 (1):\penalty0 3--15, 2006.

\bibitem[McCammon and Hollinger(2021)]{mccammon2021topological}
Seth McCammon and Geoffrey~A Hollinger.
\newblock Topological path planning for autonomous information gathering.
\newblock \emph{Autonomous Robots}, 45\penalty0 (6):\penalty0 821--842, 2021.

\bibitem[Min et~al.(2022)Min, He, Wang, and Gu]{min2022learning}
Yifei Min, Jiafan He, Tianhao Wang, and Quanquan Gu.
\newblock {Learning Stochastic Shortest Path with Linear Function
  Approximation}.
\newblock In Kamalika Chaudhuri, Stefanie Jegelka, Le~Song, Csaba Szepesvari,
  Gang Niu, and Sivan Sabato, editors, \emph{Proceedings of the 39th
  International Conference on Machine Learning}, volume 162 of
  \emph{Proceedings of Machine Learning Research}, pages 15584--15629. PMLR,
  17--23 Jul 2022.

\bibitem[{National Urban Security Technology Laboratory}(2024)]{NUSTL2024}
{National Urban Security Technology Laboratory}.
\newblock {Blue Unmanned Aircraft Systems for First Responders}.
\newblock
  \url{https://www.dhs.gov/sites/default/files/2024-03/24_03_1_st_blueuasfocusgroupreport.pdf},
  2024.

\bibitem[Newman(2018)]{newman2018networks}
Mark Newman.
\newblock \emph{Networks}.
\newblock Oxford University Press, Oxford, UK, 2nd edition, 2018.

\bibitem[Nikolova and Karger(2008)]{nikolova2008route}
Evdokia Nikolova and David~R. Karger.
\newblock {Route Planning under Uncertainty: The Canadian Traveller Problem}.
\newblock In \emph{Proceedings of the Twenty-Third AAAI Conference on
  Artificial Intelligence}, pages 969--974, Cambridge, MA, USA, 2008.
  Association for the Advancement of Artificial Intelligence, AAAI Press.

\bibitem[Powell(2019)]{powell2019unified}
Warren~B Powell.
\newblock A unified framework for stochastic optimization.
\newblock \emph{European Journal of Operational Research}, 275\penalty0
  (3):\penalty0 795--821, 2019.

\bibitem[Powell(2022)]{powell2021reinforcement}
Warren~B. Powell.
\newblock \emph{{Reinforcement Learning and Stochastic Optimization: A Unified
  Framework for Sequential Decisions}}.
\newblock John Wiley \& Sons, Inc., Hoboken, NJ, USA, 2022.

\bibitem[Powell et~al.(1995)Powell, Jaillet, and Odoni]{powell1995stochastic}
Warren~B Powell, Patrick Jaillet, and Amedeo Odoni.
\newblock Stochastic and dynamic networks and routing.
\newblock \emph{Handbooks in operations research and management science},
  8:\penalty0 141--295, 1995.

\bibitem[Ryzhov and Powell(2011)]{ryzhov2011information}
Ilya~O Ryzhov and Warren~B Powell.
\newblock {Information Collection on a Graph}.
\newblock \emph{Operations Research}, 59\penalty0 (1):\penalty0 188--201, 2011.

\bibitem[Ryzhov and Powell(2012)]{ryzhov2012information}
Ilya~O Ryzhov and Warren~B Powell.
\newblock {Information Collection for Linear Programs with Uncertain Objective
  Coefficients}.
\newblock \emph{SIAM Journal on Optimization}, 22\penalty0 (4):\penalty0
  1344--1368, 2012.

\bibitem[Secomandi(2001)]{secomandi2001rollout}
Nicola Secomandi.
\newblock A rollout policy for the vehicle routing problem with stochastic
  demands.
\newblock \emph{Operations Research}, 49\penalty0 (5):\penalty0 796--802, 2001.

\bibitem[Shahriari et~al.(2016)Shahriari, Swersky, Wang, Adams, and
  de~Freitas]{shahriari2015taking}
Bobak Shahriari, Kevin Swersky, Ziyu Wang, Ryan~P. Adams, and Nando de~Freitas.
\newblock {Taking the Human Out of the Loop: A Review of Bayesian
  Optimization}.
\newblock \emph{Proceedings of the IEEE}, 104\penalty0 (1):\penalty0 148--175,
  2016.

\bibitem[Shiri and Salman(2019)]{shiri2019randomized}
Davood Shiri and F~Sibel Salman.
\newblock {On the randomized online strategies for the k-Canadian traveler
  problem}.
\newblock \emph{Journal of Combinatorial Optimization}, 38\penalty0
  (1):\penalty0 254--267, 2019.

\bibitem[Soeffker et~al.(2022)Soeffker, Ulmer, and
  Mattfeld]{soeffker2022stochastic}
Ninja Soeffker, Marlin~W Ulmer, and Dirk~C Mattfeld.
\newblock Stochastic dynamic vehicle routing in the light of prescriptive
  analytics: A review.
\newblock \emph{European Journal of Operational Research}, 298\penalty0
  (3):\penalty0 801--820, 2022.

\bibitem[Suh et~al.(2017)Suh, Gong, and Oh]{suh2017fast}
Junghun Suh, Joonsig Gong, and Songhwai Oh.
\newblock Fast sampling-based cost-aware path planning with nonmyopic
  extensions using cross entropy.
\newblock \emph{IEEE Transactions on Robotics}, 33\penalty0 (6):\penalty0
  1313--1326, 2017.

\bibitem[Summers(2021)]{summers2021friction}
Gary~J. Summers.
\newblock {Friction and Decision Rules in Portfolio Decision Analysis}.
\newblock \emph{Decision Analysis}, 18\penalty0 (2):\penalty0 101--120, 2021.

\bibitem[Taha(2013)]{taha2013operations}
Hamdy~A Taha.
\newblock \emph{Operations research: an introduction}.
\newblock Pearson Education India, 2013.

\bibitem[Thul and Powell(2023)]{thul2023information}
Lawrence Thul and Warren~B Powell.
\newblock An information-collecting drone management problem for wildfire
  mitigation.
\newblock \emph{arXiv preprint arXiv:2301.07013}, 2023.

\bibitem[Zhang et~al.(2022)Zhang, Hartline, and Dimoulas]{zhang2022karp}
Chenhao Zhang, Jason~D Hartline, and Christos Dimoulas.
\newblock Karp: a language for np reductions.
\newblock In \emph{Proceedings of the 43rd ACM SIGPLAN International Conference
  on Programming Language Design and Implementation}, pages 762--776, 2022.

\end{thebibliography}

\end{document}